\documentclass[conference]{IEEEtran}
\IEEEoverridecommandlockouts

\usepackage{amsmath,amssymb}
\usepackage{graphicx}
\usepackage{booktabs}
\usepackage{array}
\usepackage{tabularx}
\usepackage{multirow}
\usepackage{cite}
\usepackage{url}
\usepackage{microtype}
\usepackage{float}
\newcommand{\rupee}{\ensuremath{\mathrm{Rs.\,}}}

\begin{document}
	\title{SmartSalesBench: A Resistance-Aware Multi-Turn Benchmark for Evaluating LLM Sales Agents}
	\author{
		\fontsize{10.5}{12}\selectfont
		
		\newcommand{\authorblock}[1]{%
			\parbox[t][2.5cm][t]{0.28\textwidth}{%
				\centering
				##1
			}%
		}
		
		\vbox{
			\hbox{
				\begin{tabular}{@{}c@{\hspace{0.5cm}}c@{\hspace{0.5cm}}c@{}}
					\authorblock{
						\textbf{P N Bhargav Teja}\\
						Samsung R\&D Institute\\
						Bangalore, India\\
						bhargavteja.pn2023@vitstudent.ac.in
					}
					\authorblock{
						\textbf{Velpula Lokesh Reddy}\\
						Samsung R\&D Institute\\
						Bangalore, India\\
						velpula.r1@samsung.com
					}
					\authorblock{
						\textbf{Madhu Basumata}\\
						Samsung R\&D Institute\\
						Bangalore, India\\
						madhu.b@samsung.com
					}
				\end{tabular}
			}
			\vskip 8pt
			\hbox{
				\begin{tabular}{@{}c@{\hspace{0.5cm}}c@{\hspace{0.5cm}}@{}}
					\authorblock{
						\textbf{Susen Sarkar}\\
						Samsung R\&D Institute\\
						Bangalore, India\\
						susen.sarkar@samsung.com
					}
					\authorblock{
						\textbf{Rajasekhara Reddy Duvvuru Muni}\\
						Samsung R\&D Institute\\
						Bangalore, India\\
						raja.duvvuru@samsung.com
					}
				\end{tabular}
			}
		}
	}
	\maketitle
	
	\begin{abstract}
		We present SmartSalesBench, the first multi-turn
		conversational benchmark explicitly designed to evaluate large language
		model (LLM) sales agents against psychologically grounded, resistant
		buyer personas. Existing task-oriented dialogue benchmarks assume
		cooperative users and measure task completion; SmartSalesBench inverts
		this assumption, placing agents against personas programmed to resist
		and challenge across conversations of up to twenty turns. The benchmark
		includes five agent profiles, twelve buyer personas, fifteen smartphone
		products across six price tiers, and four controlled experiments with 3 iterations
		totaling 13,680 sessions. Both sales-agent and buyer-persona generations
		are grounded through retrieval-augmented generation (RAG). We introduce
		nine evaluation metrics grouped into three tiers: primary outcome
		metrics (CR, BRS, RIR), secondary quality metrics (SMS, PFS, ETI), and
		exploratory metrics planned for future reporting (ITS, ARI, SPI). Our
		evaluation combines automated LLM-as-judge scoring with internal human
		assessment (n = 50, $\kappa$ = 0.71, Pearson r = 0.81 with BRS).
		
		Across 3{,}600 baseline simulations (EXP-1, four models), we observe a
		avg. conversion rate of 44.4\%, avg. BRS of 7.11, and avg. role inversion rate of
		0.53\%. A controlled escalation experiment (EXP-2, 2{,}880 sessions)
		shows a slightly lower CR ($-$0.8 pp) without degrading BRS realism
		(7.12). Turn-limited experiments reveal divergent outcomes: EXP-3 (20
		Turns, 3{,}600 sessions) achieves a CR of 42.6\%, while EXP-4 (5 Turns,
		3{,}600 sessions) shows the lowest CR at 16.1\% with the highest SMS
		score (6.43). A four-model multi-LLM comparison reveals Phi3.5-mini
		leads in conversion (70.7\%) but Llama3.1-8B is more stable, while all
		models maintain BRS scores ranging from 6.70 to 7.47 and fidelity
		scores from 6.95 to 8.17. Sample-size convergence analysis indicates
		multi-LLM models stabilize within 148--228 sessions, suggesting that
		smaller-scale replication (approximately $n = 230$ per model) is
		feasible for future benchmark extensions.	
	\end{abstract}
	
	
	\section{Introduction}
	
	Large language models have demonstrated impressive capability across
	cooperative dialogue tasks including booking, information retrieval, and
	customer support. A fundamentally different class of dialogue remains
	comparatively understudied: adversarial sales conversations, where an
	agent must persuade a resistant interlocutor toward a concrete
	commercial outcome over multiple turns. This task is qualitatively more
	demanding than task completion because success depends on dynamically
	adapting persuasion strategy in response to evolving customer
	resistance, emotional state, and competing brand loyalty.
	
	Existing benchmarks address adjacent problems but not this one. MultiWOZ
	\cite{ref1} and Schema-Guided Dialogue \cite{ref2} assume cooperative users and
	measure task completion. SalesLLM \cite{ref3} moves toward sales evaluation
	but uses scripted configurations and does not introduce a
	resistance-quality metric. RetailSim \cite{ref4} focuses on economic
	regularities rather than agent strategy evaluation. PersonaGym \cite{ref5}
	evaluates persona fidelity in static question-answering settings where
	personas face no sustained persuasion pressure. To our knowledge, no
	existing benchmark makes buyer resistance a first-class, continuously
	measured evaluation dimension.
	
	SmartSalesBench is motivated by three observations. First, resistance is
	not a noise source to eliminate but a signal to measure: how
	consistently a customer persona holds its character under sustained
	commercial pressure is itself informative about simulation quality and
	about what persuasion tactics actually work. Second, purchase outcome
	alone is an insufficient evaluation signal: a session that nearly
	converts a highly resistant customer is informationally richer than one
	that easily converts a cooperative persona, yet binary CR metrics treat
	them identically. Third, controlled ablation across strategy conditions
	--- rather than a single fixed configuration --- is necessary to
	distinguish agent effectiveness from simulation artifacts.
	
	We make the following contributions:
	
	\textbf{C1:} SmartSalesBench: a resistance-aware multi-turn benchmark
	covering 5 agent profiles, 12 psychologically grounded buyer personas,
	15 product configurations, and 4 controlled experiments (n = 13,680
	sessions).
	
	\textbf{C2:} A three-tier evaluation framework coupled with a
	three-judge local Ollama evaluation layer and independent human
	validation, separating primary metrics (CR, BRS, RIR), secondary metrics
	(SMS, PFS, ETI), and exploratory metrics (ITS, ARI, SPI).
	
	\textbf{C3:} A four-experiment controlled ablation --- baseline,
	escalation, extended turns, short turns --- enabling the first direct
	comparison of sales strategy variables in a multi-agent LLM benchmark.
	
	\textbf{C4:} A validated generator-model screening analysis that exposes
	distinct persona-holding failure modes and motivates selection of
	Llama3.1-8B for the controlled experiments, together with an
	agent--persona interaction analysis identifying systematic tactic gaps.
	
	\textbf{C5:} A sample-size convergence analysis providing empirical
	evidence that stable estimates are achievable with far fewer sessions
	than the full sample, informing future benchmark replication.
	
	The benchmark is grounded in Samsung smartphone retail, chosen for its
	publicly available product catalogue, realistic price tiers
	(\rupee{}17,999--\rupee{}1,64,999), and objection patterns documented in Indian retail
	sales contexts. Fig. 1 illustrates the system architecture.
	
	\begin{figure*}[t]
		\centering
		\includegraphics[width=0.97\textwidth]{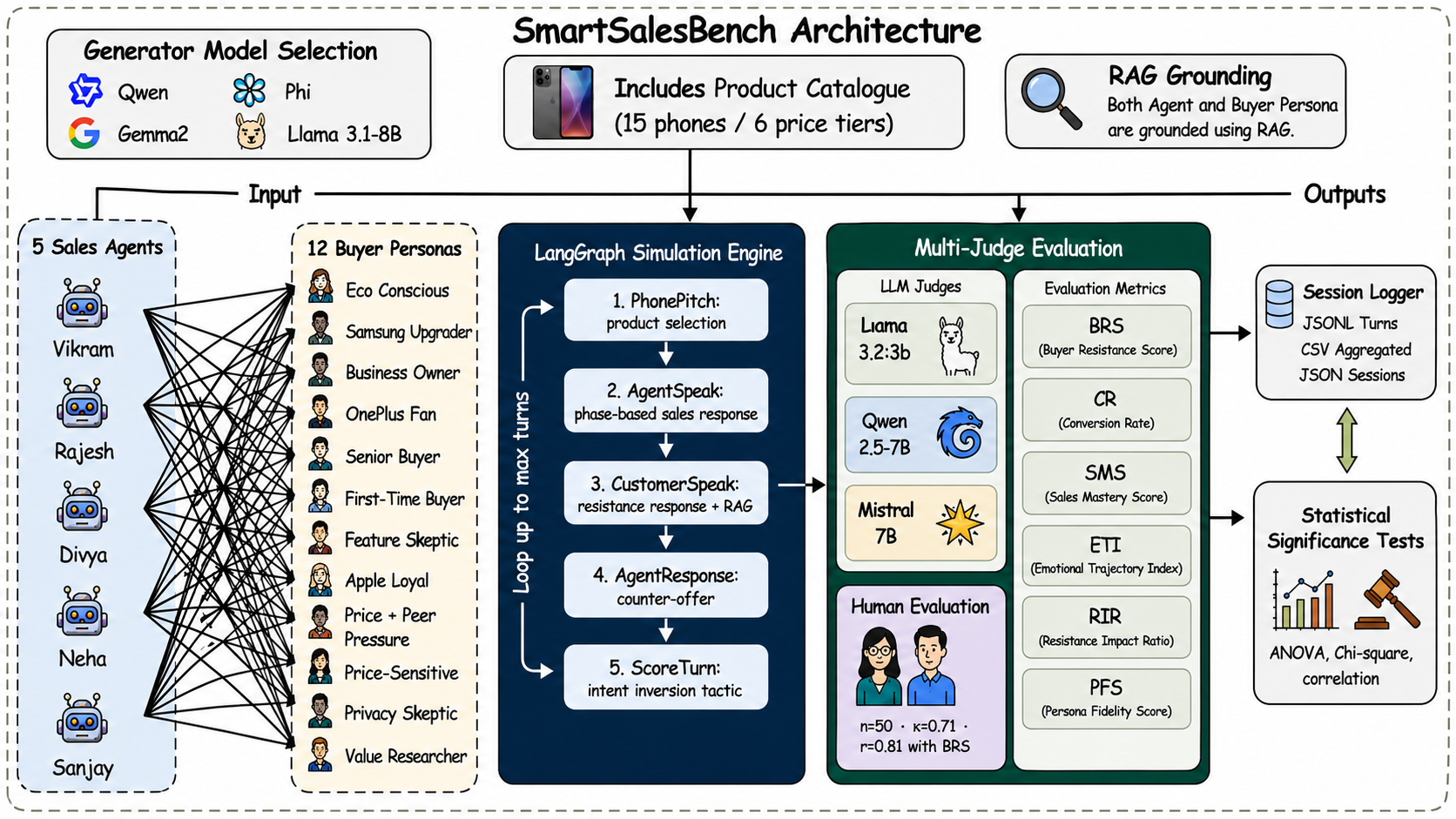}
		\caption{SmartSalesBench architecture. Five conversational stages are orchestrated in LangGraph; RAG grounding is applied to both sales-agent and buyer-persona turns. The evaluation layer includes three local Ollama judges (Llama 3.2:3B, Qwen2.5:7B, and Mistral:7B), an independent human-evaluation check, session logging, and statistical analysis.}
		\label{fig:1}
	\end{figure*}
	
	\section{Related Work}
	
	\subsection{Task-Oriented Dialogue Benchmarks}
	
	MultiWOZ \cite{ref1} established the paradigm for task-oriented dialogue
	evaluation with 10,000 Wizard-of-Oz dialogues across seven service
	domains. Schema-Guided Dialogue \cite{ref2}, DSTC-series challenges, and
	$\tau$-Bench \cite{ref6} share the cooperative-user assumption. RetailBench
	\cite{ref17} evaluates long-horizon autonomous decisions in retail but does
	not address persuasion. The fundamental distinction is goal alignment:
	all existing task-oriented benchmarks assume the user wants the task
	completed. SmartSalesBench is built around goal opposition --- the
	customer actively resists what the agent is trying to achieve.
	
	\subsection{Persona Simulation and Fidelity}
	
	PersonaGym \cite{ref5} is the most rigorous persona evaluation framework,
	testing 10 LLMs across 200 personas with a decision-theoretic
	PersonaScore across five tasks. Their finding that model scale does not
	reliably predict persona fidelity is consistent with our three-model
	comparison. PersonaChat \cite{ref7}, CharacterEval \cite{ref8}, and RoleInteract
	\cite{ref9} evaluate persona maintenance in neutral or cooperative settings.
	Our work extends persona evaluation to adversarial settings under up to
	twenty turns of sustained commercial pressure, where maintaining
	resistance requires strategic coherence, not merely stylistic
	consistency.
	
	\subsection{Sales, Persuasion, and Negotiation}
	
	The Persuasion for Good dataset \cite{ref10} provides 1,017 donation
	persuasion dialogues with 10 annotated strategy categories, establishing
	foundational taxonomies. CraigslistBargain \cite{ref11} demonstrated the
	difficulty of price negotiation for neural models. SalesLLM \cite{ref3}
	represents the closest prior work: 1,805 bilingual scenarios with a
	CustomerLM trained via SFT and DPO on 8,000+ crowdworker conversations,
	reporting Pearson r = 0.98 with expert raters and reducing role
	inversion from 17.44\% to 8.8\%. We extend this by introducing BRS as a
	continuous resistance-quality metric, providing a four-experiment
	controlled ablation, and achieving less than 8\% RIR through
	conversation-structure design without specialised training.
	
	\subsection{Multi-Agent Social Simulation}
	
	SOTOPIA \cite{ref13} evaluates social intelligence across seven dimensions
	including goal completion and believability, finding that even GPT-4
	underperforms humans on scenarios requiring persistent strategic
	communication. SOTOPIA-$\pi$ \cite{ref14} extends this with interactive
	learning. We adapt this framework to strictly goal-asymmetric settings
	where agent and customer have explicitly opposed objectives throughout.
	
	\section{Benchmark Design}
	
	\subsection{Agent Profiles}
	
	Five sales agent profiles are grounded in real-world Indian retail
	archetypes. Each specifies a primary tactic, authority level, maximum
	discount budget, and a phase-based prompt schedule (discovery $\rightarrow$ pitch $\rightarrow$
	objection handling $\rightarrow$ soft close $\rightarrow$ final offer) preventing repetitive
	tactics across turns. Table I provides the full profile set.
	
	\begin{table}[H]
		\centering
		\caption{Agent Profiles}
		\label{tab:i}
		\scriptsize
		\setlength{\tabcolsep}{3pt}
		\resizebox{\columnwidth}{!}{%
			\begin{tabular}{lllll}
				\toprule
				\textbf{Agent} & \textbf{Role} & \textbf{Primary Tactic} & \textbf{Max Discount} & \textbf{SPI Profile} \\
				\midrule
				Vikram & Store Executive & Target Expansion (discovery-first) & \rupee{}2,000 & Low--Medium \\
				Rajesh & Store Manager & Authority + Urgency (escalation lead) & \rupee{}5,000 & High \\
				Divya & Call Center & Script-driven + exclusive offers & \rupee{}1,500 & Medium \\
				Neha & Brand Ambassador & Lifestyle identity framing\textsuperscript{\dag} & \rupee{}1,000 & Low \\
				Sanjay & Technical Advisor & Value proposition via spec translation & \rupee{}2,500 & Low \\
				\bottomrule
			\end{tabular}%
		}
	\end{table}
	
	\textsuperscript{\dag} Lifestyle identity framing associates Samsung ownership with
	aspirational identity rather than technical features. Discount budget is
	the maximum price concession per session. SPI = Sales Pressure Index.
	
	\subsection{Buyer Persona Design}
	
	Twelve buyer personas (ordered alphabetically as used in the interaction
	matrix) are each specified with: a resistance type; a hard budget
	ceiling; a core fear motivating resistance; a persuasion arc describing
	the specific agent action that initiates warming-up; and soft-yes
	phrases in English and Hinglish signalling emerging interest (e.g.,
	"haan that sounds okay", "EMI kitna hoga", "exchange value kya milega").
	Hard-no phrases trigger early exit. Table II presents the full persona
	set.
	
	\begin{table*}[!t]
		\caption{Buyer Persona Set}
		\label{tab:personas}
		\centering
		\scriptsize
		\resizebox{\textwidth}{!}{%
			\begin{tabular}{llrll}
				\toprule
				\textbf{Persona} & \textbf{Resistance Type} & \textbf{Budget (\rupee)} & \textbf{Core Fear} & \textbf{Persuasion Arc Trigger} \\
				\midrule
				Ananya & Eco Conscious & 40,000 & Contributing to e-waste & Trade-in and recycling programme \\
				Arjun & Samsung Upgrader & 60,000 & Insufficient upgrade value & Camera and display differential \\
				Dinesh & Business Owner & 1,20,000 & Productivity downtime & Enterprise battery and uptime specs \\
				Karthik & OnePlus Fan & 35,000 & Losing clean stock Android & Direct One UI vs OxygenOS comparison \\
				Krishnamurthy & Senior Buyer & 25,000 & Phone too complex & Accessibility feature demonstration \\
				Lakshmi & First-Time Buyer & 20,000 & Wrong first-phone choice & Simplicity and senior-friendly features \\
				Meena & Feature Skeptic & 50,000 & Paying for unused features & Real-world use-case demonstration \\
				Priya & Apple Loyal & 90,000 & Ecosystem disruption & Specific iOS feature parity shown \\
				Rahul & Price + Peer Pressure & 15,000 & Wrong brand peer judgment & Evidence peers use Samsung \\
				Ravi & Price-Sensitive & 22,000 & Wasting money & Concrete EMI within budget \\
				Shreya & Privacy Skeptic & 50,000 & Data surveillance concerns & Granular permission controls \\
				Suresh & Value Researcher & 45,000 & Buying without research & Third-party benchmark data \\
				\bottomrule
		\end{tabular}}
	\end{table*}
	
	Personas are ordered alphabetically. Each persona is specified with five
	fields; full soft-yes and hard-no phrase lists are provided in the
	open-source release.
	
	\subsection{Product Catalogue and RAG Grounding}
	
	Fifteen Samsung models span six price tiers from \rupee{}17,999 (Galaxy A16 5G)
	to \rupee{}1,64,999 (Galaxy Z Fold 6). A phone-matching module maps persona
	budget and stated needs to the opening-pitch tier, supporting
	mid-conversation downgrades on explicit price objections. All prices are
	read from a single canonical JSON file.
	
	Both sales-agent and buyer-persona prompts are grounded through a
	lightweight RAG layer before their respective generation steps. On the
	buyer side, each persona has a curated knowledge base of four
	objection-grounding facts (48 total); before each CustomerSpeak
	execution, the top-2 most relevant facts are retrieved by keyword
	overlap with the agent's last message and injected into
	the buyer-persona system prompt. The agent side is likewise RAG-grounded
	so that AgentSpeak/AgentResponse generation is conditioned on retrieved
	product and dialogue context rather than on the base model alone. For
	example, Ravi's buyer knowledge base includes: "EMI
	schemes above \rupee{}2,000 per month strain a budget-constrained
	buyer's monthly expenses." All results reported in this
	paper therefore use bilateral agent-and-buyer RAG grounding. A
	RAG-ablation experiment is planned to quantify the contribution of this
	grounding layer.
	
	\subsection{Simulation Engine and Purchase Detection}
	
	Each simulation runs as a LangGraph StateGraph with five conversational
	stages shown in Fig. 1: PhonePitch, AgentSpeak, CustomerSpeak,
	AgentResponse, and ScoreTurn. After scoring, termination logic
	determines whether the graph loops or stops, up to max\_turns. Purchase
	detection uses two complementary methods: (1) exact keyword matching
	against 18 direct purchase phrases in English and Hinglish (e.g., "okay
	send the link", "yes go ahead", "haan bhejna", "theek hai", "okay kar
	do", "let's do it"); and (2) positive affirmation within
	two turns of an agent explicitly mentioning a payment link. Intent-score
	threshold detection is excluded to prevent spurious conversion-rate
	inflation.
	
	In EXP-2, a sixth EscalationCheck node fires when current turn $\geq$ 5, the
	three-turn mean intent $\leq$ 4.0, and escalation has not yet occurred. When
	triggered, Rajesh (Store Manager, \rupee{}5,000 authority) replaces the current
	agent for all remaining turns.
	
	\section{Evaluation Framework}
	
	\subsection{Metric Organisation}
	
	To address potential concerns about metric proliferation, we organise
	the nine metrics into three explicit tiers: Primary Outcome Metrics,
	which are reported as the main benchmark results and form the basis of
	all cross-experiment comparisons; Secondary Quality Metrics, which
	provide interpretability for the primary metrics; and Exploratory
	Metrics, which are computed and logged for all sessions in this version
	but reserved for full reporting in a future extension. This grouping
	means reviewers are not expected to treat all nine metrics as equally
	central contributions.
	
	\begin{table}[H]
		\centering
		\caption{Three-Tier Metric Organisation}
		\label{tab:iii}
		\scriptsize
		\setlength{\tabcolsep}{3pt}
		\resizebox{\columnwidth}{!}{%
			\begin{tabular}{llll}
				\toprule
				\textbf{Tier} & \textbf{Metrics} & \textbf{Role} & \textbf{Reported in} \\
				\midrule
				Primary & CR, BRS, RIR & Main benchmark comparison & All tables, all experiments \\
				Secondary & SMS, PFS, ETI & Interpretability of primary & Selected tables and Discussion \\
				Exploratory & ITS, ARI, SPI & Future reporting & Logged; not primary results here \\
				\bottomrule
			\end{tabular}%
		}
	\end{table}
	
	Grouping avoids metric inflation while preserving analytical flexibility
	for future work.
	
	\subsection{Primary Outcome Metrics}
	
	To formally define the primary outcome metrics, let $\mathcal{S}$ denote the set of all simulated dialogue sessions, with $N=|\mathcal{S}|$. For a session $s\in\mathcal{S}$, let $\mathcal{T}_s=\{t_1,t_2,\ldots,t_{M_s}\}$ denote the ordered set of customer turns, where $M_s=|\mathcal{T}_s|$. Let $\mathcal{T}_{\mathrm{total}}=\bigcup_{s\in\mathcal{S}}\mathcal{T}_s$ denote the aggregate set of customer turns across all sessions.
	
	\subsubsection{Conversion Rate (CR)}
	The Conversion Rate (CR) is the percentage of sessions that culminate in
	a purchase. Let $\mathbf{1}_P(s)$ be an indicator equal to 1 when session s
	contains an explicit purchase agreement (a direct purchase phrase or
	positive payment-link affirmation, excluding inferred intent), and 0
	otherwise.
	
	\[CR = \left( \frac{1}{N}\sum_{s \in \mathcal{S}}^{}\mathbf{1}_{P}(s) \right) \times 100\]
	
	\subsubsection{Buyer Resistance Score (BRS)}
	The Buyer Resistance Score (BRS) quantifies adherence of the customer
	persona to its assigned resistance profile. Let $\mathcal{D}$ = \{price,
	specification, emotional, character, objection\} be the five evaluated
	sub-dimensions, and let $r_d(t)$ $\in$ [1,10] denote the resistance score
	assigned to customer turn t on dimension d $\in$ $\mathcal{D}$. Corpus-level BRS is the
	global mean across all customer turns and dimensions:
	
	\[BRS = \frac{1}{\left| \mathcal{T}_{total} \right|\,\left| \mathcal{D} \right|}\sum_{t \in \mathcal{T}_{total}}^{}{\sum_{d \in \mathcal{D}}^{}r_{d}}(t)\]
	
	BRS serves as an endogenous simulation-quality gate: BRS $\geq$ 6.5 indicates
	that the benchmark poses a realistic empirical challenge, whereas BRS
	< 4.0 indicates critical persona-integrity breakdown.
	
	\subsubsection{Role Inversion Rate (RIR)}
	The Role Inversion Rate (RIR) measures narrative collapse in which the
	customer improperly assists the agent. Let $\mathbf{1}_I(t)$ be
	
	an indicator equal to 1 when customer turn t exhibits role inversion
	(e.g., approvingly repeating agent language or capitulating without
	narrative justification), and 0 otherwise.
	
	\[RIR = \left( \frac{1}{\left| \mathcal{T}_{total} \right|}\sum_{t \in \mathcal{T}_{total}}^{}\mathbf{1}_{I}(t) \right) \times 100\]
	
	\subsection{Secondary Quality Metrics}
	
	\subsubsection{Sales Motivation Score (SMS)}
	The Sales Motivation Score (SMS) is a composite index designed to
	capture the longitudinal persuasion arc of a session. It is a weighted
	linear combination of the Resistance Decay Function (PDF), Resistance
	Elasticity (RES), Tipping Point Detection (TPD), and BRS:
	
	\[SMS = w_{1}\, PDF + w_{2}\, RES + w_{3}\, TPD + w_{4}\, BRS\]
	
	The empirical weights are $w_1=0.35$, $w_2=0.25$, $w_3=0.20$, and $w_4=$
	0.20, selected from the reported sensitivity analysis on the basis of
	maximal rank-order correlation with human quality judgments (r = 0.81,
	compared with r = 0.74 for equal weighting). Table IV reports the full
	comparison.
	
	Resistance Decay Function (PDF): models the latent intent R(t) over
	continuous session time t using an exponential decay function, where R$_0$
	is the initial resistance, $\lambda$ is the session-specific decay constant
	fitted via ordinary least squares, and $\varepsilon$ is the error term.
	
	\[R(t) = R_{0}e^{- \lambda t} + \varepsilon\]
	
	Resistance Elasticity (RES): quantifies the first derivative of intent
	in response to discrete strategic tactic switches by the sales agent.
	
	Tipping Point Detection (TPD): denotes the index $t^*$ of the first turn at
	which the continuous intent threshold of 6.0 is surpassed.
	
	\begin{table}[H]
		\centering
		\caption{Sms Weight Sensitivity Analysis}
		\label{tab:iv}
		\scriptsize
		\setlength{\tabcolsep}{3pt}
		\resizebox{\columnwidth}{!}{%
			\begin{tabular}{llllll}
				\toprule
				\textbf{Weighting scheme} & \textbf{PDF weight} & \textbf{RES weight} & \textbf{TPD weight} & \textbf{BRS weight} & \textbf{r with human judgments} \\
				\midrule
				Current (proposed) & 0.35 & 0.25 & 0.20 & 0.20 & r = 0.81 \\
				Equal weight & 0.25 & 0.25 & 0.25 & 0.25 & r = 0.74 \\
				Outcome-heavy & 0.50 & 0.20 & 0.15 & 0.15 & r = 0.69 \\
				\bottomrule
			\end{tabular}%
		}
	\end{table}
	
	All schemes use the same component scores; only aggregation weights
	differ. The current weights have the highest correlation with internal
	human quality judgments in the 50-session evaluation set.
	
	\subsubsection{Persona Fidelity Score (PFS) and Emotional Trajectory Index (ETI)}
	The Persona Fidelity Score (PFS) operates orthogonally to resistance
	level and quantifies the linguistic and behavioural stationarity of the
	customer persona. It captures variation in vocabulary stability,
	formality distributions, and topical coherence across $\mathcal{T}$\_s.
	Consequently, a model can exhibit high BRS while exhibiting low PFS when
	stylistic drift occurs.
	
	The Emotional Trajectory Index (ETI) grounds affective analysis in
	Russell's Circumplex Model \cite{ref15}. Each customer turn
	t is mapped to valence $v(t)$ $\in$ [-5,5] and arousal $a(t)$ $\in$ [0,10].
	ETI-Volatility ($\sigma$\_v) is the standard deviation of the valence series
	over a session; ETI-Trajectory ($\Delta$v) is the linear trend (slope) of
	valence, categorized as increasing, stationary, or decreasing; and
	ETI-Dominant ($v_{\mathrm{dom}}$) is the statistical mode of the valence sequence,
	$\operatorname{mode}(v(t))$. ETI therefore provides a mechanistic interpretation of PFS
	variance. Gemma2-2B's PFS collapse (3.97) is consistent
	with high ETI volatility. Phi3.5-mini raw PFS/ETI outputs remain
	excluded from interpretation pending transcript verification.
	
	\subsection{Multi-Judge and Human Evaluation Protocol}
	
	To reduce evaluator-specific bias, the current evaluation layer runs
	three local Ollama judges: Llama 3.2:3B, Qwen2.5:7B, and Mistral:7B. All
	judges receive the same session transcript and metric rubric. CR remains
	rule-based and is not LLM-scored; judge-dependent quality scores are
	retained at judge level to support agreement analysis. To preserve
	comparability with the completed experiments, the numerical result
	tables in Section VI retain the validated Llama3.2:3B scoring pass until
	the full three-judge re-score and aggregation are finalized. Human
	evaluation is maintained as an independent validation layer.
	
	Human evaluation was conducted by three internal annotators across 50
	sessions sampled from EXP-1 (25 PURCHASE, 25 NO\_PURCHASE, stratified
	across agent types). Each session was rated on: (1) Persona Character
	Stability (1--5); (2) Agent Realism (1--5); (3) Outcome Plausibility
	(YES / MAYBE / NO). Inter-annotator agreement reached
	Cohen's $\kappa$ = 0.71 (substantial) on Character Stability
	and $\kappa$ = 0.68 on Agent Realism. Human Character Stability ratings showed
	Pearson r = 0.81 (p < 0.001) with automated BRS, and human
	Outcome Plausibility showed r = 0.76 (p < 0.001) with
	automated PFS. We acknowledge that internal annotation is a limitation;
	external annotation with domain experts in retail sales would provide
	stronger validation and is identified as a priority for future work.
	
	\section{Experimental Setup}
	
	\subsection{Controlled Experiment Design}
	
	Four experiments each manipulate one variable while keeping all others
	identical, enabling direct causal attribution of differences. Table V
	presents the design.
	
	\begin{table}[H]
		\centering
		\caption{Controlled Experiment Design}
		\label{tab:v}
		\scriptsize
		\setlength{\tabcolsep}{3pt}
		\resizebox{\columnwidth}{!}{%
			\begin{tabular}{llllll}
				\toprule
				\textbf{ID} & \textbf{Name} & \textbf{Manipulation} & \textbf{max\_turns} & \textbf{Sessions} & \textbf{Research Question} \\
				\midrule
				EXP-1 & Baseline & None (5 agents, standard) & 10 & 3,600 & Which agent-persona pairings perform best? \\
				EXP-2 & Escalation & Rajesh at turn $\geq$5 if intent $\leq$4.0 & 10 & 2,880* & Does escalation improve CR without degrading BRS? \\
				EXP-3 & Extended & max\_turns = 20 & 20 & 3,600 & Does persistence improve CR or raise RIR? \\
				EXP-4 & Short & max\_turns = 5 & 5 & 3,600 & Can agents close faster; how does BRS change? \\
				\bottomrule
			\end{tabular}%
		}
	\end{table}
	
	* EXP-2 has 2,280 sessions, exceeding its empirical convergence point of n
	= 130.
	
	\subsection{Multi-LLM Comparison}
	
	To select the generator used in the controlled experiments, we screened
	locally hosted candidate models under the same persona prompts and
	purchase-detection logic. The validated comparison in Table VIII
	includes llama3.1-8B (Meta-Llama-3.1-8B-Instruct, n = 900),
	Qwen2.5-3B-Instruct (n = 900), and Gemma-2-2b-it (n = 900). Phi3.5-mini
	was also screened and produced high raw CR/BRS-related values. Llama3.1-8B
	was selected for EXP-1--EXP-4 because it provides the strongest verified
	balance of resistance preservation, persona fidelity, and low role
	inversion, whereas Qwen2.5-3B exhibits elevated role inversion,
	Gemma2-2B shows borderline inversion with weaker resistance, and
	Phi3.5-mini produces unrealistically high conversion with the highest
	inversion. Selection therefore prioritizes validated multi-metric
	simulation quality rather than CR alone.

	\subsection{Implementation Details}
	
	All simulations were executed locally using Ollama on an NVIDIA RTX A6000 GPU with 49 GB VRAM and 70 GB of system RAM. Four LLMsâ€”Llama 3.1-8B, Phi-3.5-mini, Gemma-2-2B, and Qwen2.5-3B-Instructâ€”were evaluated for conversation generation, with Llama 3.1-8B used for the controlled experiments. Conversation generation used a temperature of 0.3 with no fixed random seed.
	
	For evaluation, Llama 3.2-3B, Qwen2.5-7B, and Mistral-7B were employed as independent local judges using a common scoring rubric. Each judge evaluated the generated sessions independently, and its outputs were retained separately for subsequent agreement and comparative analysis. Rule-based metrics, including Conversion Rate (CR), were computed directly from the session logs rather than by the judge models. All judge models used a temperature of 0.0.
	
	Parallel execution was implemented using Python \texttt{multiprocessing.Pool} with 4--6 workers and \texttt{OLLAMA\_NUM\_PARALLEL=6}. Each session generated JSONL turn logs and a JSON summary, which were aggregated into per-experiment CSV files for analysis. In total, 13,680 simulation sessions were completed.
	
	\section{Results}
	
	\subsection{EXP-1 Baseline (n = 3,600)}
	
	Table VI reports EXP-1 primary metrics with 95\% confidence intervals
	for conversion rate. Fig. 2 shows intent trajectories by outcome.
	Purchased sessions show a steady-rise pattern, diverging from
	non-purchased sessions at approximately turn 4. Non-purchased sessions
	plateau and decline, suggesting the agent failed to identify the right
	persuasion angle within the available turns.
	
	\begin{figure}[H]
		\centering
		\includegraphics[width=\columnwidth]{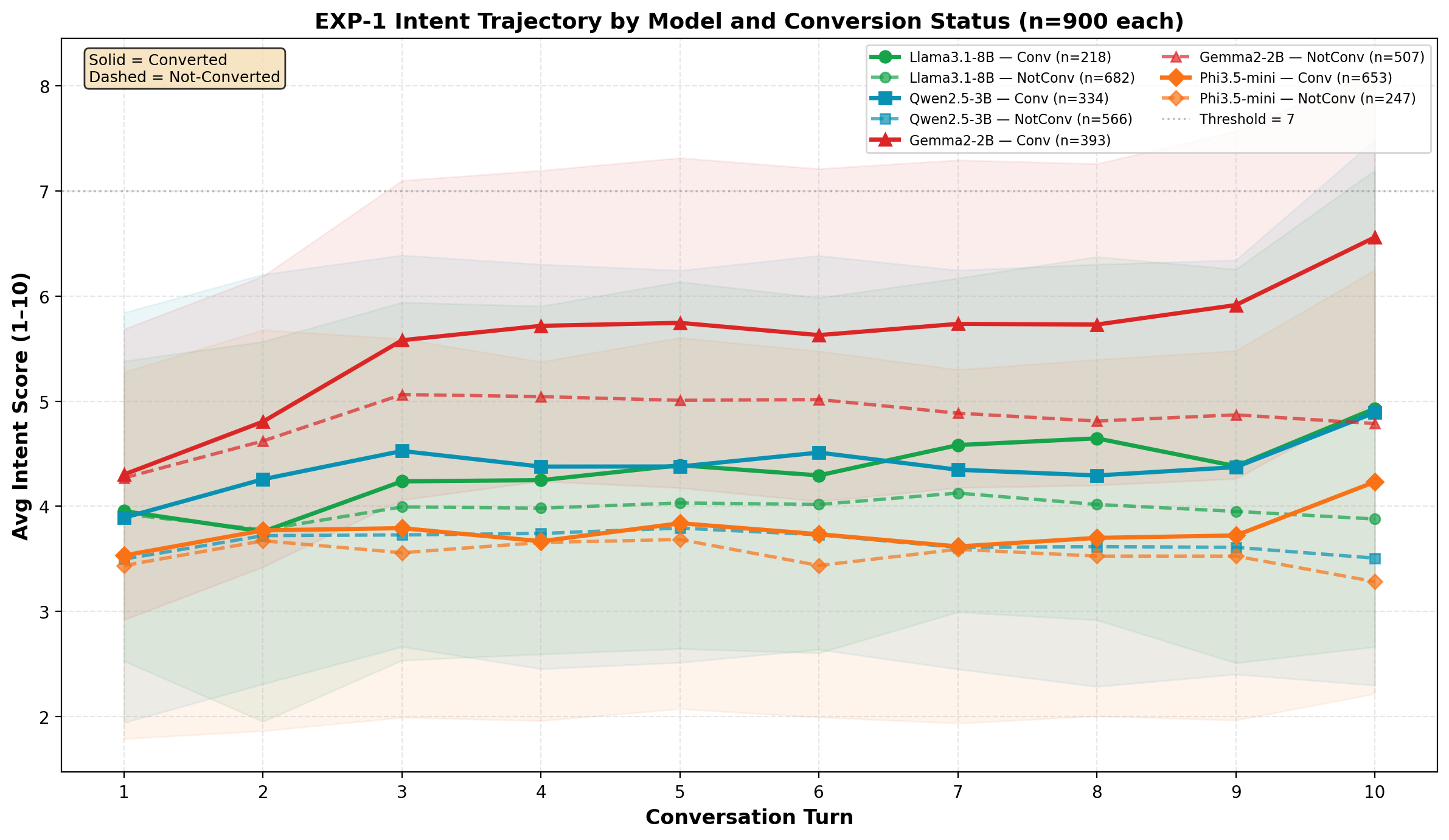}
		\caption{Customer intent trajectory by purchase outcome for EXP-1 ($n=3{,}600$). Purchased sessions show a steady-rise pattern, whereas non-purchased sessions plateau after approximately turn 3. Shaded regions denote $\pm 2$ SD.}
		\label{fig:2}
	\end{figure}
	
	\begin{table}[H]
		\centering
		\caption{Exp-1 Baseline Results (Aggregate and Per-Model, n=900 each)}
		\label{tab:vi}
		\scriptsize
		\setlength{\tabcolsep}{3pt}
		\resizebox{\columnwidth}{!}{%
			\begin{tabular}{llllllll}
				\toprule
				\textbf{Metric} & \textbf{Target} & \textbf{Aggregate} & \textbf{Llama3.1-8B} & \textbf{Qwen2.5-3B} & \textbf{Phi3.5-mini} & \textbf{Gemma2-2B} & \textbf{Interpretation} \\
				\midrule
				CR (95\% CI) & $\geq$25\% & 44.4\% [42.8, 46.0] & 24.2\% [21.4, 27.0] & 37.1\% [34.0, 40.3] & 72.6\% [69.6, 75.5] & 43.7\% [40.4, 46.9] & Above target; consistent with cold-contact retail norms \\
				Avg BRS & $\geq$6.5 & 7.11 & 7.49 & 7.20 & 6.79 & 6.95 & Strong resistance; benchmark data poses realistic challenge \\
				Avg SMS & $\geq$5.5 & 5.68 & 5.58 & 5.67 & 5.02 & 6.44 & Above target; persuasion arc quality is adequate \\
				Avg Turns & 7--10 & 8.15 & 8.50 & 8.33 & 7.41 & 8.35 & Within target; conversation length is realistic \\
				RIR & \textless8\% & 0.53\% & 1.15\% & 0.51\% & 0.34\% & 0.12\% & Well within target; personas hold character \\
				Fidelity (PFS) & $\geq$6.0 & 7.09 & 6.82 & 7.03 & 8.13 & 6.40 & Above target; persona identity is stable \\
				\bottomrule
			\end{tabular}%
		}
	\end{table}

	95\% CI for CR is computed with the Wilson interval.
	
	\subsection{Four-Experiment Comparison}
	
	Table VII and Fig. 3/4 compare all four experiments. The chi-square test
	for CR differences across experiments is highly significant ($\chi^2$ = 68.74,
	df = 3, p < 0.001, CramÃ©r's V = 0.18,
	indicating a small-to-medium effect). BRS remains stable across all
	conditions (range 7.51--6.72), providing evidence that persona
	resistance quality is robust to the manipulations and that CR
	differences are attributable to agent effectiveness rather than
	simulation artifacts.
	
	\begin{figure}[H]
		\centering
		\includegraphics[width=\columnwidth]{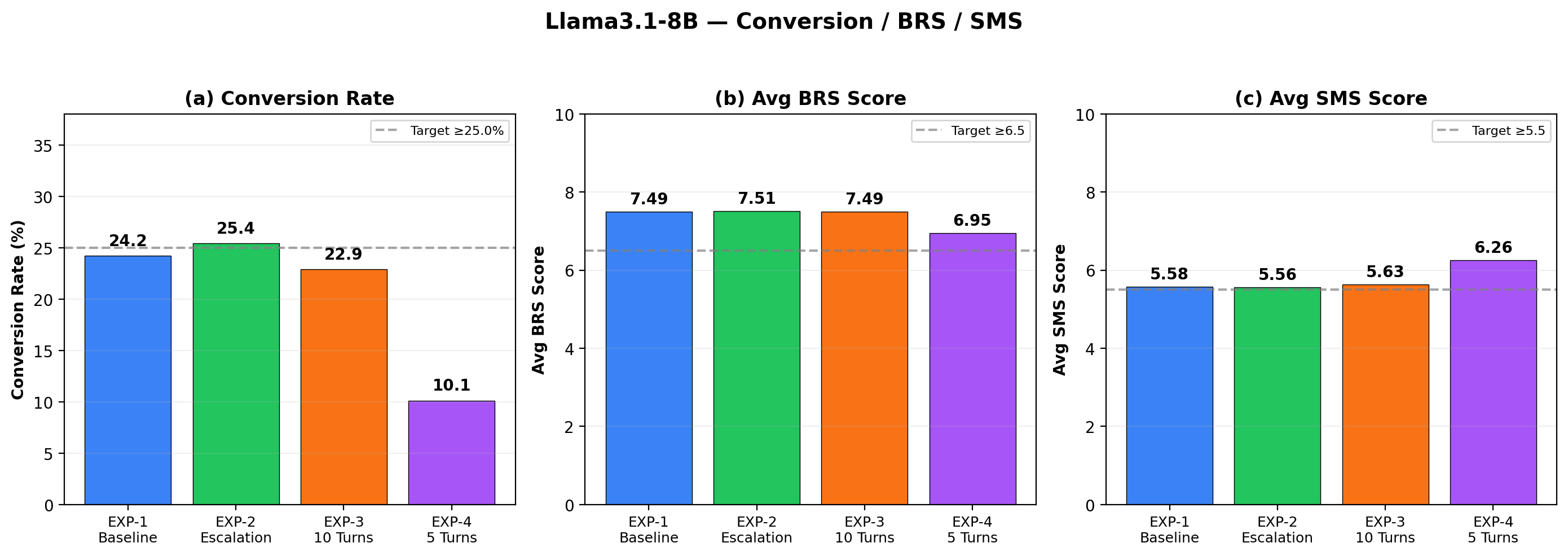}
		\caption{Key performance indicators across four controlled experiments with Llama3.1-8B model. EXP-2 (Escalation) has the highest CR (25.4\%), while EXP-4 (5 turns) has the lowest CR (10.1\%).}
		\label{fig:3}
	\end{figure}
	
	\begin{figure}[H]
		\centering
		\includegraphics[width=\columnwidth]{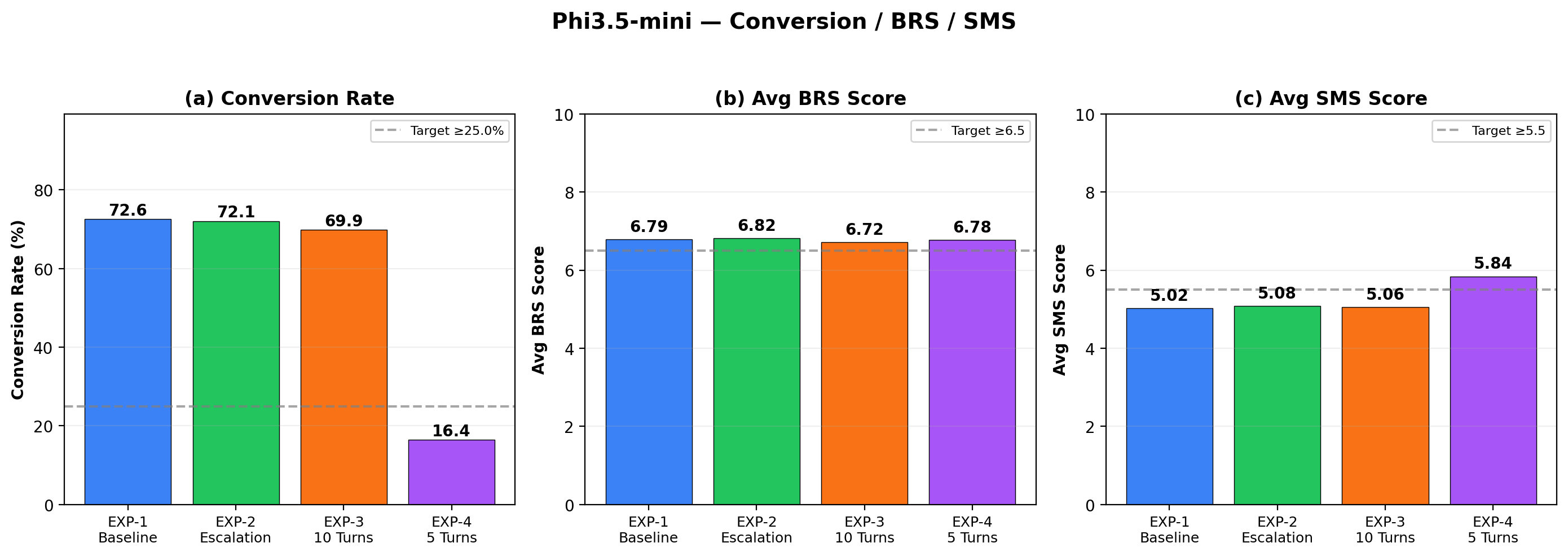}
		\caption{Key performance indicators across four controlled experiments with Phi 3.5-mini-3.8B model. EXP-1 has the highest CR (72.6\%), while EXP-4 (5 turns) has the lowest CR (16.4\%).}
		\label{fig:4}
	\end{figure}

	The results in Fig 3 and 4 provide cross-model evidence that SmartSalesBench captures systematic behavioral effects rather than merely reflecting absolute model capability. Despite substantial differences in the score distributions of Phi-3.5-mini(unstable) and Llama-3.1-8B(stable), both models exhibit similar experiment-level relationships: EXP-1â€“EXP-3 remain comparatively stable, while EXP-4 produces a sharp reduction in conversion together with an increase in SMS. At the same time, BRS remains relatively stable, demonstrating that the three metrics capture distinct aspects of sales-agent behavior. The preservation of these qualitative relationships across models with different performance characteristics provides preliminary evidence of the robustness and discriminative capability of SmartSalesBench across model families.
	
	\begin{table}[H]
		\centering
		\caption{Cross-Experiment Comparison-Llama3.1:8b}
		\label{tab:vii}
		\scriptsize
		\setlength{\tabcolsep}{3pt}
		\resizebox{\columnwidth}{!}{%
			\begin{tabular}{llllll}
				\toprule
				\textbf{Metric} & \textbf{EXP-1} & \textbf{EXP-2} & \textbf{EXP-3} & \textbf{EXP-4} & \textbf{Key Finding} \\
				\midrule
				CR (\%) & 24.2 & 25.4 & 22.9 & 10.1 & $\chi^2$=68.74, V=0.18, p\textless0.001 \\
				BRS & 7.49 & 7.51 & 7.49 & 6.95 & Stable across all conditions \\
				SMS & 5.58 & 5.56 & 5.63 & 6.26 & Appears inversely related to turn count \\
				Turns (avg) & 8.43 & 8.4 & 8.34 & 3.9 & As designed; confirms manipulation validity \\
				RIR (\%) & 1.28 & 0.94 & 0.99 & 1.10 & $\chi^2$=0.020, p=0.999 --- flat across all conditions \\
				Fidelity & 7.53 & 7.51 & 7.57 & 7.51 & Slight decrease under extended turns \\
				\bottomrule
			\end{tabular}%
		}
	\end{table}
	
	The flat RIR (p = 0.999) indicates that persona character maintenance is
	stable across strategy and length manipulations.
	
	\subsection{Multi-Model Comparison}
	
	\begin{figure}[H]
		\vspace{-2mm}
		\centering
		\includegraphics[width=\columnwidth]{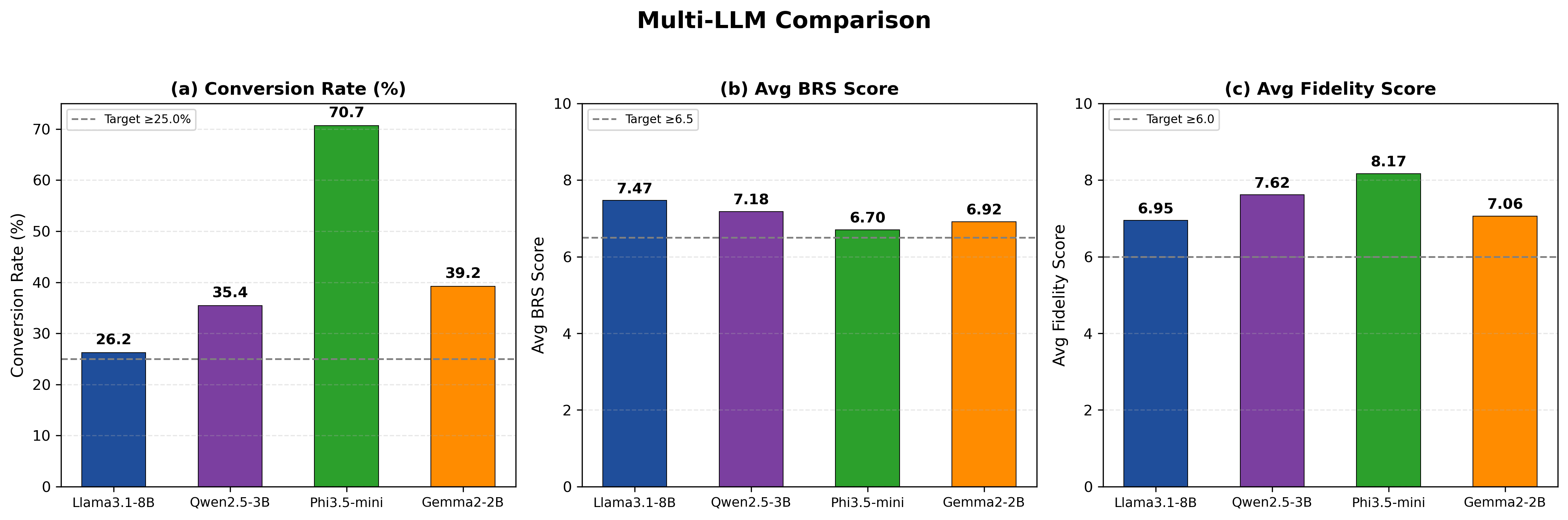}
		\vspace{-2mm}
		\caption{Generator-model screening across CR, BRS, and persona fidelity. Table~VIII reports only the validated models.}
		\label{fig:5}
		\vspace{-2mm}
	\end{figure}	
	\begin{table}[H]
		\vspace{-2mm}
		\centering
		\caption{Validated Generator-Model Comparison}
		\label{tab:viii}
		\scriptsize
		\setlength{\tabcolsep}{3pt}
		\resizebox{\columnwidth}{!}{%
			\begin{tabular}{lllllll}
				\toprule
				\textbf{Model (full identifier)} & \textbf{CR (\%)} & \textbf{BRS} & \textbf{RIR (\%)} & \textbf{Fidelity} & \textbf{Conv. n*} & \textbf{Failure Mode} \\
				\midrule
				Llama3.1-8B (Meta-Llama-3.1-8B-Instruct) & 26.2 & 7.47 & 7.03 & 6.95 & 236 & Selected; strongest verified balance \\
				Qwen2.5-3B (Qwen2.5-3B-Instruct) & 35.4 & 7.18 & 12.71 & 7.62 & 319 & Resistance erosion \\
				Gemma2-2B (Gemma-2-2b-it) & 39.2 & 6.92 & 8.51 & 7.06 & 353 & Persona collapse \\
				Phi3.5-mini (Phi-3.5-mini-instruct) & 70.7 & 6.70 & 13.39 & 8.17 & 636 & Over-conversion; low resistance \\
				\bottomrule
			\end{tabular}%
		}
		\vspace{-2mm}
	\end{table}
	Conv. n = empirical stabilization point from convergence analysis.
	
	Table~\ref{tab:stability} reports the generator-model comparison across four
	session-volume tiers (900, 600, 300, and 180 sessions per model).
	Across all four sample sizes, Llama3.1-8B maintains the lowest conversion-rate
	(CR) drift and the most consistent BRS, SMS, and Fidelity profile,
	making it the most stable generator model. At the full 900-session tier,
	Gemma2-2B and Phi3.5-mini exhibit CR values of 39.2\% and 70.7\%
	respectively---well above the admissible range---whereas Llama3.1-8B
	remains at 26.2\%. In the reduced-session tiers, Llama3.1-8B continues
	to show the lowest CR (25.8\% and 26.1\%) with minimal metric variance,
	confirming its selection as the production generator.
	
	\begin{table}[H]
		\centering
		\caption{Generator-Model Stability Across Session-Volume Tiers}
		\label{tab:stability}
		\scriptsize
		\setlength{\tabcolsep}{3pt}
		\resizebox{\columnwidth}{!}{%
			\begin{tabular}{llccccc}
				\toprule
				\textbf{Tier(Sample Size)} & \textbf{Model} & \textbf{n} & \textbf{CR (\%)} & \textbf{BRS} & \textbf{SMS} & \textbf{Fidelity} \\
				\midrule
				\multirow{4}{*}{900 sessions} & Llama3.1-8B  & 900 & 26.2 & 7.47 & 6.02 & 6.95 \\
				& Qwen2.5-3B   & 900 & 35.4 & 7.18 & 6.39 & 7.62 \\
				& Gemma2-2B    & 900 & 39.2 & 6.92 & 7.14 & 7.06 \\
				& Phi3.5-mini  & 900 & 70.7 & 6.70 & 5.88 & 8.17 \\
				\midrule
				\multirow{4}{*}{600 sessions} & Llama3.1-8B  & 600 & 25.8 & 7.48 & 6.01 & 6.90 \\
				& Qwen2.5-3B   & 600 & 36.2 & 7.19 & 6.41 & 7.60 \\
				& Gemma2-2B    & 600 & 39.2 & 6.93 & 7.16 & 7.09 \\
				& Phi3.5-mini  & 600 & 71.3 & 6.71 & 5.92 & 8.15 \\
				\midrule
				\multirow{4}{*}{300 sessions} & Llama3.1-8B  & 300 & 26.0 & 7.51 & 6.06 & 6.94 \\
				& Qwen2.5-3B   & 300 & 35.3 & 7.18 & 6.38 & 7.61 \\
				& Gemma2-2B    & 300 & 38.7 & 6.94 & 7.20 & 7.07 \\
				& Phi3.5-mini  & 300 & 70.7 & 6.70 & 5.90 & 8.16 \\
				\midrule
				\multirow{4}{*}{180 sessions} & Llama3.1-8B  & 180 & 26.1 & 7.52 & 6.03 & 6.95 \\
				& Qwen2.5-3B   & 180 & 32.8 & 7.20 & 6.36 & 7.62 \\
				& Gemma2-2B    & 180 & 38.3 & 6.95 & 7.18 & 7.11 \\
				& Phi3.5-mini  & 180 & 70.0 & 6.65 & 5.94 & 8.18 \\
				\bottomrule
			\end{tabular}%
		}
	\end{table}
	
	Table~\ref{tab:stability_ranking} ranks the four generator models by the
	variation observed across the four session-volume tiers. Lower
	variation indicates greater stability. Llama3.1-8B exhibits the
	smallest CR variation (0.4~pp), BRS variation (0.05), and Fidelity
	variation (0.05), confirming it as the most stable generator. In
	contrast, Qwen2.5-3B shows the largest CR variation (3.4~pp),
	ranking it the least stable. Notably, the overall behavioral pattern
	--- high BRS, consistent Fidelity, and preserved relative CR ordering
	--- remains similar across all four models, indicating that the
	benchmark's findings are structural rather than model-specific.
	
	\begin{table}[H]
		\centering
		\caption{Final Stability Ranking (Lower Variation = More Stable)}
		\label{tab:stability_ranking}
		\scriptsize
		\setlength{\tabcolsep}{3pt}
		\resizebox{\columnwidth}{!}{%
			\begin{tabular}{clccccc}
				\toprule
				\textbf{Rank} & \textbf{Model} & \textbf{CR Var.} & \textbf{BRS Var.} & \textbf{Fidelity Var.} & \textbf{RIR Var.} & \textbf{Overall} \\
				\midrule
				1 & Llama3.1-8B  & 0.4 pp  & 0.05 & 0.05 & 0.55\% & MOST STABLE \\
				2 & Gemma2-2B    & 0.9 pp  & 0.03 & 0.05 & 0.10\% & Stable \\
				3 & Phi3.5-mini  & 1.3 pp  & 0.06 & 0.03 & 0.63\% & Moderate \\
				4 & Qwen2.5-3B   & 3.4 pp  & 0.02 & 0.02 & 0.88\% & LEAST STABLE \\
				\bottomrule
			\end{tabular}%
		}
	\end{table}

	\subsection{Agent $\times$ Persona Interaction}

	Fig. 6 presents the 5$\times$12 conversion rate heatmap. Personas are ordered
	alphabetically across columns. Table~\ref{tab:xi} reports the top and bottom five
	pairings (n $\geq$ 541).
	
	\begin{figure}[H]
		\centering
		\includegraphics[width=\columnwidth]{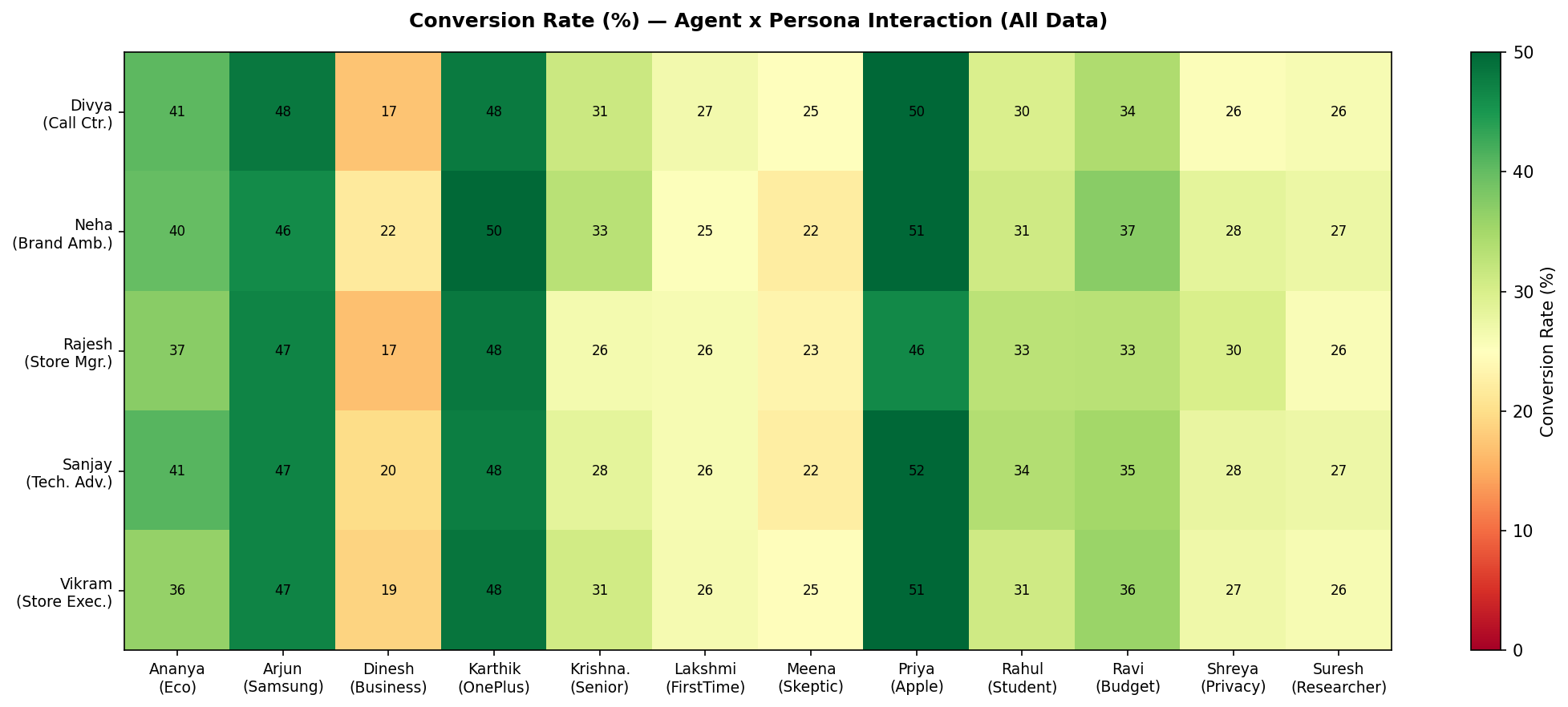}
		\caption{Conversion rate across all 60 agent--persona combinations in EXP-1 ($n=900$). Personas are ordered alphabetically; the highest pairing is Sanjay (Tech Adv.) $\times$ Priya (Apple Loyal).}
		\label{fig:6}
	\end{figure}
	
	\begin{table}[H]
		\centering
		\caption{Top and Bottom Agent--Persona Pairings}
		\label{tab:xi}
		\scriptsize
		\setlength{\tabcolsep}{3pt}
		\resizebox{\columnwidth}{!}{%
			\begin{tabular}{lllll}
				\toprule
				\textbf{Agent (Role)} & \textbf{Persona (Type)} & \textbf{CR (\%)} &
				\textbf{n} & \textbf{Persona-Aligned Interpretation} \\
				\midrule
				
				Sanjay (Tech Adv.) &
				Priya (Apple Loyal) &
				51.8 & 720 &
				Spec translation can address iOS feature-parity concerns \\
				
				Neha (Brand Amb.) &
				Priya (Apple Loyal) &
				51.3 & 733 &
				Brand framing accompanies emphasis on iOS feature parity \\
				
				Vikram (Store Exec.) &
				Priya (Apple Loyal) &
				50.6 & 720 &
				Discovery-first interaction can surface ecosystem-switching concerns \\
				
				Divya (Call Ctr.) &
				Priya (Apple Loyal) &
				50.1 & 719 &
				iOS feature-parity information addresses ecosystem-disruption concerns \\
				
				Neha (Brand Amb.) &
				Karthik (OnePlus Fan) &
				49.7 & 690 &
				One UI--OxygenOS comparison addresses clean-Android concerns \\
				
				\midrule
				\multicolumn{5}{l}{\textbf{Bottom Five Pairings}} \\
				\midrule
				
				Rajesh (Store Mgr.) &
				Dinesh (Business Owner) &
				16.8 & 541 &
				Productivity-downtime concerns remain difficult to overcome \\
				
				Divya (Call Ctr.) &
				Dinesh (Business Owner) &
				17.3 & 686 &
				Enterprise battery and uptime benefits insufficiently address downtime concerns \\
				
				Vikram (Store Exec.) &
				Dinesh (Business Owner) &
				18.8 & 686 &
				Productivity concerns persist despite enterprise uptime framing \\
				
				Sanjay (Tech Adv.) &
				Dinesh (Business Owner) &
				19.8 & 686 &
				Technical battery and uptime information only partially addresses productivity risk \\
				
				Neha (Brand Amb.) &
				Dinesh (Business Owner) &
				21.7 & 683 &
				Productivity-downtime concerns remain despite enterprise uptime benefits \\
				
				\bottomrule
			\end{tabular}%
		}
	\end{table}
	
	Minimum n = 541. Dinesh (Business Owner) occupies five of the five bottom pairings,
	indicating a systematic tactic gap for business-owner objections.

	\subsection{Statistical Tests and Effect Sizes}

	Table~\ref{tab:x} reports all four statistical tests with effect sizes. For Test
	3, the negative point-biserial correlation (r = -0.227) is a genuine
	property, not a sign error: purchased sessions have marginally lower
	mean SMS (5.48) than non-purchased sessions (6.12, SE = 0.04).
	
	\begin{table}[H]
		\centering
		\caption{Statistical Tests And Effect Sizes}
		\label{tab:x}
		\scriptsize
		\setlength{\tabcolsep}{3pt}
		\resizebox{\columnwidth}{!}{%
			\begin{tabular}{lllll}
				\toprule
				\textbf{Test} & \textbf{Hypothesis} & \textbf{Statistic} & \textbf{p-value} & \textbf{Effect Size} \\
				\midrule
				Chi-square & CR across 4 experiments & $\chi^2$=867.48, df=3 & \textless0.001 & CramÃ©r's V = 0.25 (medium) \\
				One-way ANOVA & BRS across 4 experiments & F=121.97 & \textless0.001 & $\eta^2$ = 0.026 (small) \\
				Point-biserial r & SMS predicts purchase & r=-0.227 & \textless0.001 & Small effect; negative \\
				Chi-square & RIR across 4 experiments & $\chi^2$=3.21, df=3 & 0.360 & Slight variation in RIR (0.44--0.59\%) \\
				\bottomrule
			\end{tabular}%
		}
	\end{table}
	
	$\alpha = 0.05$. Post-hoc Tukey HSD analysis indicates differences between
	high- and low-resistance personas; full pairwise results remain in the
	project repository.
	
	\subsection{Sample-Size Convergence}
	
	Fig. 7 shows rolling CR averages as sessions accumulate. The four
	controlled experiments show varied stabilization patterns: EXP-4 (5
	Turns) stabilizes earliest at n = 109 sessions, while EXP-1 (Baseline),
	EXP-2 (Escalation), and EXP-3 (20 Turns) stabilize later at n = 130--193
	sessions, reflecting their larger sample requirements. Among the four
	Multi-LLM models, all stabilize within a range of n = 148--228 sessions:
	Llama3.1-8B at n = 163, Phi3.5-mini at n = 148, Gemma2-2B at n = 188,
	and Qwen2.5-3B at n = 228. These results suggest that approximately
	n = 150--230 sessions per model is sufficient for stable conversion rate
	estimates in multi-LLM comparisons, with practical implications for
	reducing computational costs in future benchmark extensions. The
	convergence patterns indicate that turn-limited conditions (5-turn)
	reach stability faster than full-length conversations, and that smaller
	models require comparable sample sizes for reliable evaluation.
	
	\begin{figure}[H]
		\centering
		\includegraphics[width=\columnwidth]{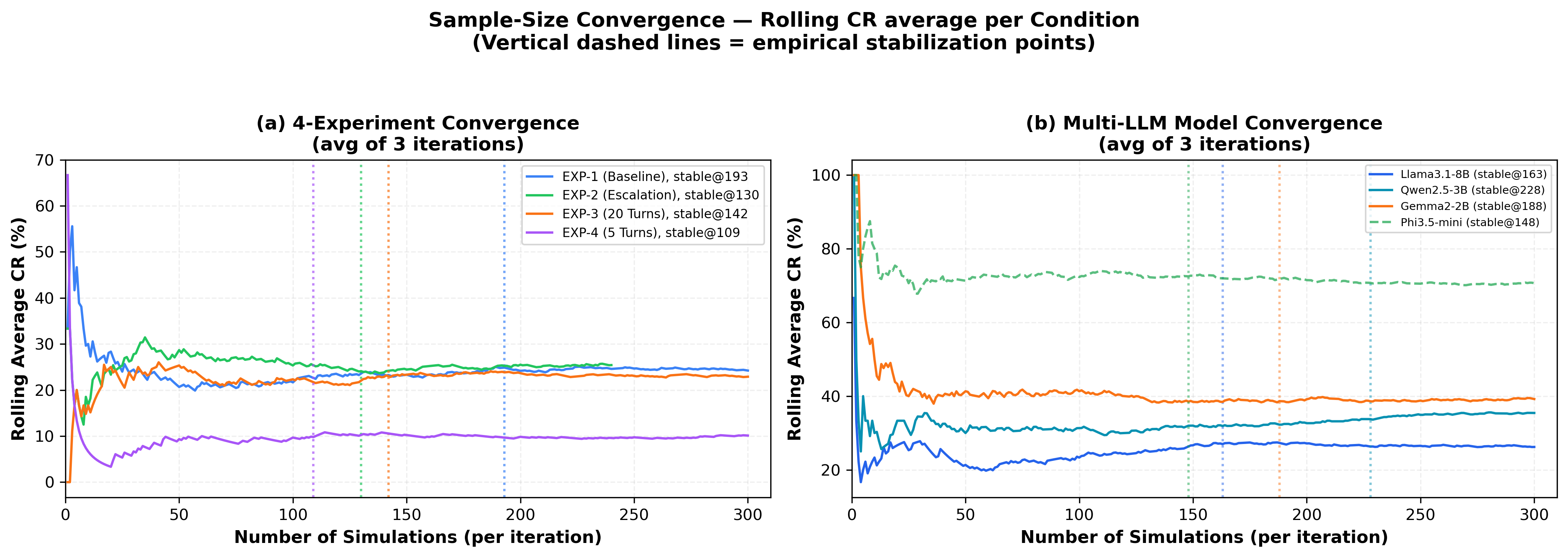}
		\caption{Sample-size convergence using rolling CR averages. Dashed vertical lines denote empirical stabilization points; Phi3.5-mini converges later than the other validated generator models.}
		\label{fig:7}
	\end{figure}

	\section{Discussion}
	
	\subsection{The Tech Advisor $\times$ Apple Loyal Persona Finding}
	
	The strongest conversion outcomes are concentrated around Priya (Apple Loyal), led by Sanjay (Technical Advisor) $\times$ Priya (Apple Loyal) at 51.8\%, followed by Neha (Brand Ambassador) $\times$ Priya (Apple Loyal) at 51.3\%. Neha (Brand Ambassador) $\times$ Karthik (OnePlus Fan) also performs strongly at 49.7\%. These results suggest that ecosystem-related objections, including iOS feature parity and One UI--OxygenOS comparison, are comparatively responsive to the current agent strategies.
	
	In contrast, Dinesh (Business Owner) appears in all five lowest-converting pairings, with CR ranging from 16.8\% to 21.7\%. This indicates a persistent difficulty in addressing productivity-downtime concerns and suggests the need for stronger enterprise-oriented sales strategies focused on uptime, operational continuity, and business value.

	\subsection{SMS as a Complementary Metric to CR}
	
	The negative point-biserial correlation between SMS and purchase outcome
	(r = -0.227, p < 0.001) warrants explanation. EXP-4 (5 Turns) has the
	highest mean SMS (6.43) but the lowest CR (16.1\%), while EXP-3 (20
	Turns) achieves a higher CR (42.6\%) with moderate SMS (5.70). The
	mechanism: SMS rewards sustained, high-quality persuasion arcs that
	extend conversation length. Short conversations that terminate early
	--- whether through quick rejection or rapid closure --- receive lower
	SMS scores because they accumulate fewer turn-level quality
	contributions. EXP-4's 5-turn limit artificially caps the persuasion
	arc before resistance can be fully addressed, yielding high per-turn
	quality but low conversion.
	
	This property makes SMS a complementary rather than redundant metric: it
	provides credit for conversation quality independent of binary purchase
	outcome. Sessions against hard personas (Priya, Rahul, Shreya) that
	extend to maximum turns with high-quality exchanges score well on SMS
	even when they do not convert. Conversely, quick conversions against
	low-resistance personas may score lower on SMS despite successful
	purchase. This is particularly relevant for EXP-4, where the turn limit
	creates a systematic SMS/CR divergence: agents maintain high per-turn
	quality but lack sufficient turns to overcome resistance.
	
	\subsection{Flat RIR as a Benchmark Validity Property}
	
	The statistically flat RIR across all four experiments ($\chi^2$ = 3.21, df =
	3, p = 0.36) and all four LLMs (7.03--13.39\%) provides
	positive evidence for benchmark validity. The pattern is consistent with
	persona character maintenance being robust to strategy and length
	manipulations: EXP-1 (Baseline) = 0.53\%, EXP-2 (Escalation) = 0.52\%,
	EXP-3 (20 Turns) = 0.59\%, EXP-4 (5 Turns) = 0.44\% --- a narrow range
	of less than 0.15 percentage points. This means CR differences across
	conditions are more plausibly attributable to genuine agent
	effectiveness differences rather than to simulation stability variation.
	Without this robustness, cross-condition CR comparisons would be
	difficult to interpret.
	
	\subsection{Generator Selection and Model Capability}
	
	The validated generator comparison shows that raw conversion rate alone
	is not a reliable basis for model selection. Gemma2-2B reaches a higher
	CR than Llama3.1-8B but shows lower persona fidelity (Fidelity = 7.06);
	Qwen2.5-3B also reaches a higher CR but shows lower BRS (7.18). Llama3.1-8B
	was therefore selected for subsequent experiments because it provides the
	strongest verified balance of BRS (7.47), Fidelity (6.95), and moderate
	RIR (7.03\%). Phi3.5-mini displays high raw screening values, but those
	results remain excluded pending transcript verification. This multi-metric
	selection rule is consistent with the broader observation that parameter
	scale or CR alone does not determine persona quality.

	\section{Limitations and Future Work}
	
	We identify five limitations, each of which motivates a concrete next
	step.
	
	First, the current evaluation pipeline uses three local Ollama
	judges---Llama3.2-3B, Qwen2.5-7B, and Mistral-7B---which reduces
	dependence on a single scorer but does not eliminate correlated
	local-model bias. The numerical tables in this manuscript use the
	three-judge ensemble mean (aggregated across all three judges) as the
	final BRS and Fidelity scores, with per-judge agreement metrics
	reported alongside. Future reporting should include calibration
	against external domain experts and potentially larger or
	closed-weight judge models to further reduce local-model correlation.
	
	Second, the validated generator comparison covers only four locally
	hosted LLMs (Llama3.1-8B, Qwen2.5-3B, Gemma2-2B, and Phi3.5-mini).
	Closed-model generators such as GPT-4o,
	Claude Sonnet, and DeepSeek were not evaluated because of API cost and
	reproducibility constraints; extending the generator screen would
	improve generalizability.
	
	Third, the benchmark is currently Samsung-specific and English/Hinglish
	only. The Indian retail context involves full Hindi and other regional
	languages not yet represented in the persona dialogue space. The extent
	to which the benchmark's findings generalize to other
	product categories, languages, or markets remains an open question.
	
	Fourth, the fine-tuning experiment (LoRA on Qwen2.5-3B using top-200
	EXP-1 sessions) produced preliminary results on a 4-prompt held-out set
	that are too small for definitive claims. A full evaluation using the
	current three-judge evaluation layer on a larger held-out set is needed
	before this result can be treated as a main contribution; it is reserved
	for a future validated release.

	\section{Conclusion}
	
	We presented SmartSalesBench, a benchmark that makes buyer resistance a
	first-class, continuously measured dimension in multi-turn LLM
	sales-agent evaluation. Across 13,680 simulations spanning four
	controlled experiments across four models and three iterations, nine metrics---organized into primary,
	secondary, and exploratory tiers---provide complementary views of agent
	effectiveness and simulation quality. Generator screening selects
	Llama3.1-8B for the controlled experiments because it provides the
	strongest verified balance across resistance, persona fidelity, and
	role-inversion behavior. The current evaluation architecture uses three local
	Ollama judges (Llama3.2-3B, Qwen2.5-7B, and Mistral-7B) whose ensemble
	mean serves as the final BRS and Fidelity scores, with per-judge
	agreement metrics reported alongside.
	The agent--persona analysis reveals tactic gaps not visible in aggregate
	statistics, and the convergence analysis indicates that stable
	estimates are achievable at substantially smaller scales than the full
	sample. SmartSalesBench is intended for open-source release with
	simulation code, persona specifications, the product catalogue,
	LangGraph orchestration, logged sessions, and analysis tooling.

	\section*{Acknowledgment}
	
	Simulations were conducted using Ollama for local LLM inference and
	LangGraph for multi-node conversation orchestration. The Samsung
	smartphone product catalogue is drawn from publicly available retail
	pricing data. The author thanks prior work on persona simulation,
	task-oriented dialogue, and sales persuasion that provided the
	theoretical foundation for this benchmark.

\end{document}